\documentclass[11pt,a4paper]{article}
\usepackage[hyperref]{acl2019}

\usepackage[utf8]{inputenc} 
\usepackage[T1]{fontenc}    
\usepackage{url}            
\usepackage{booktabs}       
\usepackage{amsfonts}       
\usepackage{nicefrac}       
\usepackage{microtype}      
\usepackage{bm}

\usepackage{times}
\usepackage{latexsym}
\usepackage{helvet}  
\usepackage{courier}  
\usepackage{graphicx}
\usepackage{subfigure}  

\usepackage{color}
\usepackage{multirow}
\usepackage{bbm}
\usepackage{algorithm,algorithmic}
\usepackage{float}
\usepackage{caption}
\usepackage{wrapfig}
\usepackage{amsmath}
\usepackage{lipsum}

\newcommand{\Imat}{{\bf I}}

\newcommand{\Kmat}[0]{{{\bf K}}}

\newcommand{\Qmat}[0]{{{\bf Q}}}

\newcommand{\Vmat}[0]{{{\bf V}}}
\newcommand{\Wmat}[0]{{{\bf W}}}
\newcommand{\Xmat}[0]{{{\bf X}}}
\newcommand{\Ymat}{{\bf Y}}

\newcommand{\bv}[0]{{\boldsymbol{b}}}

\newcommand{\ev}[0]{{\boldsymbol{e}}}

\newcommand{\hv}[0]{{\boldsymbol{h}}}

\newcommand{\sv}[0]{{\boldsymbol{s}}}

\newcommand{\vv}{\boldsymbol{v}}
\newcommand{\wv}{\boldsymbol{w}}
\newcommand{\xv}{\boldsymbol{x}}
\newcommand{\yv}{\boldsymbol{y}}
\newcommand{\zv}{\boldsymbol{z}}

\newcommand{\Sigmamat}[0]{{\boldsymbol{\Sigma}}}

\newcommand{\muv}[0]{{\boldsymbol{\mu}}}

\newcommand{\Ncal}{\mathcal{N}}

\aclfinalcopy 
\def\aclpaperid{1593} 

\title{Unsupervised Abstractive Dialogue Summarization for Tete-a-Tetes}

\author{Xinyuan Zhang$^1$, Ruiyi Zhang$^2$, Manzil Zaheer$^3$, Amr Ahmed$^3$\\
	$^1$ASAPP Inc.\\
	$^2$Duke University\\
	$^3$Google Research\\
	\texttt{xzhang@asapp.com}}

\date{}

\begin{document}
	\maketitle
	\begin{abstract}
		High-quality dialogue-summary paired data is expensive to produce and domain-sensitive, making abstractive dialogue summarization a challenging task.
		In this work, we propose the first unsupervised abstractive dialogue summarization model for tete-a-tetes (SuTaT).
		Unlike standard text summarization, a dialogue summarization method should consider the multi-speaker scenario where the speakers have different roles, goals, and language styles.
		In a tete-a-tete, such as a customer-agent conversation, SuTaT aims to summarize for each speaker by modeling the customer utterances and the agent utterances separately while retaining their correlations.
		SuTaT consists of a conditional generative module and two unsupervised summarization modules.
		The conditional generative module contains two encoders and two decoders in a variational autoencoder framework where the dependencies between two latent spaces are captured.
		With the same encoders and decoders, two unsupervised summarization modules equipped with sentence-level self-attention mechanisms generate summaries without using any annotations.
		Experimental results show that SuTaT is superior on unsupervised dialogue summarization for both automatic and human evaluations, and is capable of dialogue classification and single-turn conversation generation.
	\end{abstract}
	
	\section{Introduction}
	
	Tete-a-tetes, conversations between two participants, have been widely studied as an importance component of dialogue analysis.
	For instance, tete-a-tetes between customers and agents contain information for contact centers to understand the problems of customers and improve the solutions by agents.
	However, it is time-consuming for others to track the progress by going through long and sometimes uninformative utterances.
	Automatically summarizing a tete-a-tete into a shorter version while retaining its main points can save a vast amount of human resources and has a number of potential real-world applications.
	
	Summarization models can be categorized into two classes: extractive and abstractive.
	Extractive methods select sentences or phrases from the input text, while abstractive methods attempt to generate novel expressions which requires an advanced ability to paraphrase and condense information.
	Despite being easier, extractive summarization is often not preferred in dialogues for its limited capability to capture highly dependent conversation histories and produce coherent discourses.
	Therefore, abstractively summarizing dialogues has attracted recent research interest \cite{goo2018abstractive,pan2018dial2desc,yuan2019abstractive,liu2019automatic}.
	
	\begin{table}
		\centering
		\def\arraystretch{1.0}
		\setlength{\tabcolsep}{3pt}
		\small
		\begin{tabular} {p{0.6in} p{2.2in}}
			\toprule
			\textbf{Customer:} & I am looking for the Hamilton Lodge in Cambridge. \\ 
			\textbf{Agent:} & Sure, it is at 156 Chesterton Road, postcode cb41da. \\
			\textbf{Customer:} & Please book it for 2 people, 5 nights beginning on Tuesday. \\ 
			\textbf{Agent:} & Done. Your reference number is qnvdz4rt. \\
			\textbf{Customer:} & Thank you, I will be there on Tuesday! \\ 
			\textbf{Agent:} & Is there anything more I can assist you with today? \\
			\textbf{Customer:} & Thank you! That's everything I needed. \\ 
			\textbf{Agent:} & You are welcome. Any time. \\
			\midrule
			\textbf{Customer Summary:} & i would like to book a hotel in cambridge on tuesday . \\ 
			\textbf{Agent Summary:} & i have booked you a hotel . the reference number is qnvdz4rt . can i help you with anything else ?\\
			\bottomrule
		\end{tabular}
		\caption{An example of SuTaT generated summaries.}
		\label{tab:example} 
	\end{table}
	
	However, existing abstractive dialogue summarization approaches fail to address two main problems.
	First, a dialogue is carried out between multiple speakers and each of them has different roles, goals, and language styles.
	Taking the example of a contact center, customers aim to propose problems while agents aim to provide solutions, which leads them to have different semantic contents and choices of vocabularies.
	Most existing methods process dialogue utterances as in text summarization without accommodating the multi-speaker scenario.
	Second, high-quality annotated data is not readily available in the dialogue summarization domain and can be very expensive to produce.
	Topic descriptions or instructions are commonly used as gold references which are too general and lack any information about the speakers.
	Moreover, some methods use auxiliary information such as dialogue acts \cite{goo2018abstractive}, semantic scaffolds \cite{yuan2019abstractive} and key point sequences \cite{liu2019automatic} to help with summarization, adding more burden on data annotation.
	To our knowledge, no previous work has focused on unsupervised deep learning for abstractive dialogue summarization.
	
	We propose SuTaT, an unsupervised abstractive dialogue summarization approach specifically for tete-a-tetes.
	In this paper, we use the example of \textit{agent} and \textit{customer} to represent the two speakers in tete-a-tetes for better understanding.
	In addition to summarization, SuTaT can also be used for dialogue classification and single-turn conversation generation.
	
	To accommodate the two-speaker scenario, SuTaT processes the utterances of a customer and an agent separately in a conditional generative module.
	Inspired by \citet{zhang2019syntax} where two latent spaces are contained in one variational autoencoder (VAE) framework, the conditional generative module includes two encoders to map a customer utterance and the corresponding agent utterance into two latent representations, and two decoders to reconstruct the utterances jointly.
	Separate encoders and decoders enables SuTaT to model the differences of language styles and vocabularies between customer utterances and agent utterances.
	The dependencies between two latent spaces are captured by making the agent latent variable conditioned on the customer latent variable.
	Compared to using two standard autoencoders that learn deterministic representations for input utterances, using the VAE-based conditional generative module to learn variational distributions gives the model more expressive capacity and more flexibility to find the correlation between two latent spaces.
	
	The same encoders and decoders from the conditional generative module are used in two unsupervised summarization modules to generate customer summaries and agent summaries.
	Divergent from MeanSum \cite{chu2018meansum} where the combined multi-document representation is simply computed by averaging the encoded input texts, SuTaT employs a setence-level self-attention mechanism \cite{vaswani2017attention} to highlight more significant utterances and neglect uninformative ones.
	We also incorporate copying factual details from the source text that has proven useful in supervised summarization \cite{see2017get}.
	Dialogue summaries are usually written in the third-person point of view, but SuTaT simplifies this problem by making the summaries consistent with the utterances in pronouns.
	Table \ref{tab:example} shows an example of SuTaT generated summaries.
	
	Experiments are conducted on two dialogue datasets: MultiWOZ \cite{budzianowski2018multiwoz} and Taskmaster \cite{byrne2019taskmaster}.
	It is assumed that we can only access utterances in the datasets without any annotations including dialogue acts, descriptions, instructions, etc.
	Both automatic and human evaluations show SuTaT outperforms other unsupervised baseline methods on dialogue summarization.
	We further show the capability of SuTaT on dialogue classification with generated summaries and single-turn conversation generation.
	
	\section{Methodology}
	\begin{figure*}
		\centering
		\includegraphics[width=6in]{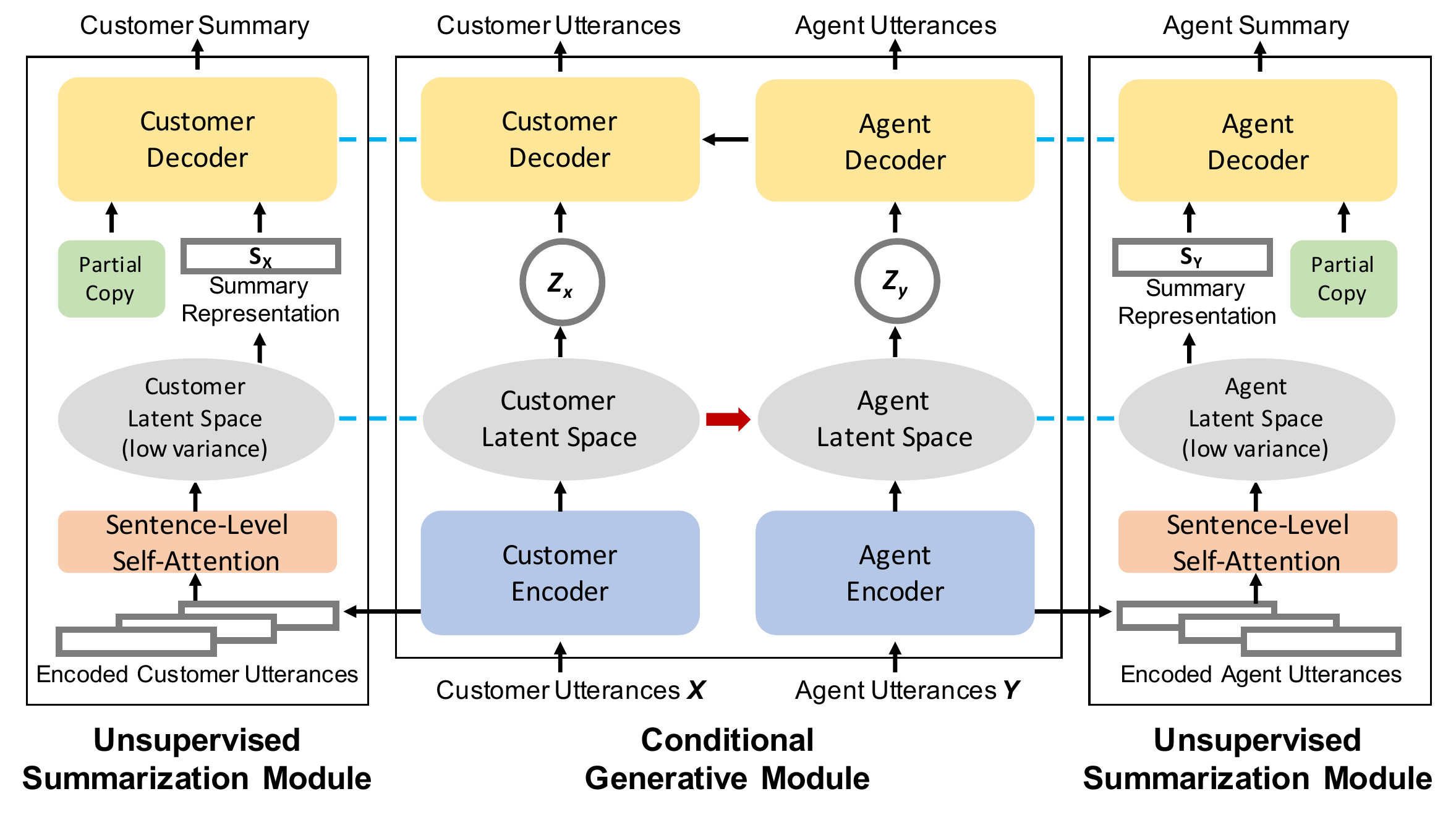}
		\caption{Block diagram of SuTaT. Architectures connected by a blue dashed line are the same. The red arrow represents the conditional relationship between two latent spaces.}
		\label{fig:diag}
	\end{figure*}
	
	SuTaT consists of a conditional generative module and two unsupervised summarization modules.
	Let $\Xmat=\{\xv_1,\cdots,\xv_n\}$ denote a set of customer utterances and $\Ymat=\{\yv_1,\cdots,\yv_n\}$ denote a set of agent utterances in the same dialogue.
	Our aim is to generate a customer summary and an agent summary for the utterances in $\Xmat$ and $\Ymat$.
	
	Figure \ref{fig:diag} shows the entire architecture of SuTaT.
	Given a customer utterance $\xv$ and its consecutive agent utterance $\yv$, the conditional generative module embeds them with two encoders and obtain latent variables $\zv_x$ and $\zv_y$ from the variational latent spaces, then reconstruct the utterances from $\zv_x$ and $\zv_y$ with two decoders.
	In the latent space, the agent latent variable is conditioned on the customer latent variable; during decoding, the generated customer utterances are conditioned on the generated agent utterances.
	This design resembles how a tete-a-tete carries out: the agent's responses and the customer's requests are dependent on each other.
	The encoded utterances of a dialogue are the inputs of the unsupervised summarization modules.
	We employ a sentence-level self-attention mechanism on the utterances embeddings to highlight the more informative ones and combine the weighted embeddings.
	A summary representation is drawn from the low-variance latent space using the combined utterance embedding, which is then decoded into a summary with the same decoder and a partial copy mechanism.
	The whole process does not require any annotations from the data.
	
	\subsection{Conditional Generative Module}
	We build the conditional generative module in a SIVAE-based framework \cite{zhang2019syntax} to capture the dependencies between two latent spaces.
	The goal of the module is to train two encoders and two decoders for customer utterances $\xv$ and agent utterances $\yv$ by maximizing the evidence lower bound
	\begin{align}\label{eq:elbo}
	&\mathcal{L}_{gen}=\mathbb{E}_{q(\zv_x|\xv)}\log p(\xv|\yv,\zv_x) - \\
	\nonumber&\operatorname{KL} [q(\zv_x | \xv)||p(\zv_x)]+\mathbb{E}_{q(\zv_y|\yv,\zv_x)} \log p(\yv | \zv_y)\\
	\nonumber&-\operatorname{KL} [ q(\zv_y | \yv, \zv_x) || p(\zv_y | \zv_x) ] \leq \log p (\xv, \yv),
	\end{align}
	where $q(\cdot)$ is the variational posterior distribution that approximates the true posterior distribution.
	The lower bound includes two reconstruction losses and two Kullback-Leibler (KL) divergences between the priors and the variational posteriors.
	By assuming priors and posteriors to be Gaussian, we can apply the reparameterization trick \cite{kingma2014auto} to compute the KL divergences in closed forms.
	$q(\zv_x|\xv)$, $q(\zv_y|\yv,\zv_x)$, $p(\xv|\yv,\zv_x)$, and $p(\yv | \zv_y)$ represent customer encoder, agent encoder, customer decoder, and agent decoder.
	
	The correlation between two latent spaces are captured by making the agent latent variable $\zv_y$ conditioned on the customer latent variable $\zv_x$.
	We define the customer prior $p(\zv_x)$ to be a standard Gaussian $\Ncal(\mathbf{0},\Imat)$. 
	The agent prior $p(\zv_y|\zv_x)$ is also a Gaussian $\Ncal(\muv,\Sigmamat)$ where the mean and the variance are functions of $\zv_x$,
	\begin{align}
	\nonumber\muv = \operatorname{MLP}_\mu(\zv_x),\ \ \ \Sigmamat = \operatorname{MLP}_\Sigma(\zv_x).
	\end{align}
	This process resembles how a tete-a-tete at contact centers carries out: the response of an agent is conditioned on what the customer says.
	
	\paragraph{Encoding}
	Given a customer utterance sequence $\xv=\{\wv_1,\cdots,\wv_t\}$, we first encode it into an utterance embedding $\ev_x$ using bidirectional LSTM \cite{graves2013hybrid} or a Transformer encoder \cite{vaswani2017attention}.
	
	The Bi-LSTM takes the hidden states $\hv_i=[\overrightarrow{\hv}_i;\overleftarrow{\hv}_i]$ as contextual representations by processing a sequence from both directions,
	\begin{align*}
	\overrightarrow{\hv}_i=\operatorname{LSTM}(\wv_i,\hv_{i-1}), \overleftarrow{\hv}_i=\operatorname{LSTM}(\wv_i,\hv_{i+1}).
	\end{align*}
	The Transformer encoder produces the contextual representations that have the same dimensions as word embeddings,
	\begin{align}
	\nonumber\{\dot{\wv}_1,\cdots,\dot{\wv}_t\} =  \operatorname{TransEnc}(\{\wv_1,\cdots,\wv_t\}).
	\end{align}
	The customer utterance embedding $\ev_x$ is obtained by averaging over the contextual representations.
	Similarly, we can obtain the agent utterance embedding $\ev_y$.
	
	The customer latent variable $\zv_x$ is first sampled from $q(\zv_x|\xv)=\mathcal{N}(\muv_x,\Sigmamat_x)$ using $\ev_x$, then the agent latent variable $\zv_y$ is sampled from $q(\zv_y|\yv,\zv_x)=\mathcal{N}(\muv_y,\Sigmamat_y)$ using $\ev_y$ and $\zv_x$.
	The Gaussian parameters $\muv_x$, $\Sigmamat_x$, $\muv_y$ and $\Sigmamat_y$ are computed with separate linear projections,
	\begin{align*}
	&\muv_x = \operatorname{Linear}_{\mu_x}(\ev_x),\muv_y = \operatorname{Linear}_{\mu_y}(\ev_y\oplus\zv_x) \\
	&\Sigmamat_x = \operatorname{Linear}_{\Sigma_x}(\ev_x), \Sigmamat_y = \operatorname{Linear}_{\Sigma_y}(\ev_y\oplus\zv_x).
	\end{align*}
	
	\paragraph{Decoding}
	We first decode $\zv_y$ into the agent utterance from the $p(\yv | \zv_y)$ using LSTM \cite{sutskever2014sequence} or a Transformer decoder \cite{vaswani2017attention}.
	The decoded sequence and the latent variable $\zv_x$ are then used in $p(\xv|\yv,\zv_x)$ to generate the customer utterance.
	
	In the LSTM decoder,
	\begin{align*}
	&\vv_y^{(i)}=\operatorname{LSTM}(\yv_{i-1},\zv_y,\vv_y^{(i-1)})\\
	&\vv_x^{(i)}=\operatorname{LSTM}(\xv_{i-1},\zv_x\oplus\yv,\vv_x^{(i-1)}).
	\end{align*}
	While in the Transformer decoder,
	\begin{align*}
	&\vv_y^{(i)}=\operatorname{TranDec}(\yv_{<i},\zv_y)\\
	&\vv_x^{(i)}=\operatorname{TranDec}(\xv_{<i},\zv_x\oplus\yv)
	\end{align*}
	where $\yv_{<i}$ and $\xv_{<i}$ are the embeddings of the previously decoded sequence.
	The decoded representations $\vv_y^{(i)}$ and $\vv_x^{(i)}$ are put in feedforward layers to compute the vocabulary distributions,
	\begin{align}\label{eq:decode}
	\nonumber&p(\yv_i|\yv_{<i},\zv_y)=\operatorname{softmax} (\vv_y^{(i)}\Wmat_y^T+\bv_y)\\
	&p(\xv_i|\xv_{<i},\zv_x,\yv)=\operatorname{softmax} (\vv_x^{(i)}\Wmat_x^T+\bv_x)
	\end{align}
	where $\Wmat_x\in\mathbb{R}^{|x|\times l}$, $\Wmat_y\in\mathbb{R}^{|y|\times l}$, $\bv_x\in\mathbb{R}^l$ and $\bv_y\in\mathbb{R}^l$ are learnable parameters.
	$|x|$ and $|y|$ are the vocabulary sizes for customer utterances and agent utterances.
	
	\subsection{Unsupervised Summarization Module}
	Given the encoded utterances of a dialogue, an unsupervised summarization module learns to generate a summary that is semantically similar to the input utterances using trained components from the conditional generative module.
	
	\paragraph{Sentence-Level Self-Attention}
	Some utterances like greetings or small talk do not contribute to the content of a dialogue.
	Therefore, we employ a sentence-level self-attention mechanism, which is built upon Multi-head attention \cite{vaswani2017attention}, to highlight the most significant utterances in a dialogue.
	
	The multi-head attention partitions the queries $\Qmat$, keys $\Kmat$, and values $\Vmat$ into $h$ heads along their dimensions $d$, and calculates $h$ scaled dot-product attention for the linear projections of the heads.
	\begin{align*}
	\operatorname{MH}&(\Qmat,\Kmat,\Vmat)=\operatorname{Concat}(head_1,\cdots,head_h)\Wmat^O\\
	&head_i=\operatorname{SDP}(\Qmat\Wmat_i^Q,\Kmat\Wmat_i^K,\Vmat\Wmat_i^V)
	\end{align*}
	where $\Wmat^O$, $\Wmat^Q$, $\Wmat^K$, and $\Wmat^V$ are trainable parameters.
	The scaled dot-product attention outputs a weighted sum of values,
	\begin{align*}
	\operatorname{SDP}(\Qmat,\Kmat,\Vmat)=\operatorname{softmax}(\frac{\Qmat\Kmat^T}{\sqrt{d}})\Vmat.
	\end{align*}
	
	In SuTaT, the sentence-level self-attention is achieved by making the queries, keys, and values all be the set of encoded agent/customer utterances of a dialogue.
	The self-attention module assigns weights on the input utterances such that more significant and informative ones have higher weights.
	The output is a weighted combined utterance embedding $\tilde{\ev}_X$ or $\tilde{\ev}_Y$ that highlights more informative utterances from the dialogue.
	
	\paragraph{Summary Generation}
	Summary representations $\sv_X$ and $\sv_Y$ are sampled from the latent spaces taking the weighted combined utterance representations $\tilde{\ev}_X$ and $\tilde{\ev}_Y$ as inputs.
	To limit the amount of novelty in the generated summary, we set the variances of the latent spaces close to zero so that $\sv_X\approx\muv_x$ and $\sv_Y\approx\muv_y$.
	$\sv_X$ and $\sv_Y$ containing key information from the dialogue are decoded into a customer summary and an agent summary using the same decoders from the conditional generative module, which makes the generated summaries similar to the utterances in pronouns and language styles.
	
	We re-encode the generated summaries into $\ev_X$ and $\ev_Y$ with the same encoders and compare them with each of the utterance embeddings using average cosine distance.
	To constrain the summaries to be semantically close to input utterances, the summarization modules are trained by maximizing a similarity loss,
	\begin{align}\label{eq:sum}
	\mathcal{L}_{sum}=\frac{1}{n}\sum_{i=1}^{n}(\operatorname{d}(\ev_X,\ev_x^{(i)})+\operatorname{d}(\ev_Y,\ev_y^{(i)})),
	\end{align}
	where $\operatorname{d}$ denotes the cosine distance.
	
	However, the summarization modules are prone to produce inaccurate factual details.
	We design a simple but effective partial copy mechanism that employs some extractive summarization tricks to address this problem.
	We automatically make a list of factual information from the data such as dates, locations, names, and numbers.
	Whenever the decoder predicts a word from the factual information list, the copy mechanism replaces it with a word containing factual information from the input utterances.
	If there are multiple factual information words in the dialogue, the one with the highest predictive possibility will be chosen.
	Note that this partial copy mechanism does not need to be trained and is not activated during training.
	
	\subsection{Training Process}
	The objective function we optimize is the weighted sum of the reconstruction loss in Equation \ref{eq:elbo} and the similarity loss in Equation \ref{eq:sum},
	\begin{align}\label{eq:obj}
	\mathcal{L}=\alpha\mathcal{L}_{gen}+(1-\alpha)\mathcal{L}_{sum},
	\end{align}
	where $\alpha$ controls the weights of two objectives.
	
	SuTaT involves re-encoding the generated agent utterance to help with generating the customer utterance in Equation \ref{eq:decode} and re-encoding the generated summary to compare with utterance embeddings in Equation \ref{eq:sum}.
	Directly sampling from the multinomial distribution with $\operatorname{argmax}$ is a non-differentiable operation, so we use the soft-argmax trick \cite{chen2019improving} to approximate the deterministic sampling scheme,
	\begin{align}\label{eq:soft}
	\yv_i=\operatorname{softmax}(\vv_Y^{(i)}/\tau),
	\end{align}
	where $\tau\in(0,1)$ is the annealing parameter.
	
	Adam \cite{kingma2014adam} is adopted for stochastic optimization to jointly train all model parameters by maximizing Equation \ref{eq:obj}. 
	In each step, Adam samples a mini-batch of dialogues and then updates the parameters \cite{zhang2018diffusion}.
	
	\section{Related Works}
	\paragraph{Dialogue Summarization}
	Early dialogue summarization works mainly focus on extractively summarizing using statistical machine learning methods \cite{galley2006skip,xie2008evaluating,wang2013domain}.
	Abstractive dialogue summarization has been recently explored due to the success of sequence-to-sequence neural networks.
	\citet{pan2018dial2desc} propose an enhanced interaction dialogue encoder and a transformer-pointer decoder to summarize dialogues.
	\citet{li2019keep} summarize multi-modal meetings on another encoder-decoder structure.
	Some approaches design additional mechanisms in a neural summarization model to leverage auxiliary information such as dialogue acts \cite{goo2018abstractive}, key point sequences \cite{liu2019automatic}, and semantic scaffolds \cite{yuan2019abstractive}.
	However, these supervised methods can only use concise topic descriptions or instructions as gold references while high-quality annotated dialogue summaries are not readily available.
	
	\paragraph{Unsupervised Summarization}
	Many extractive summarization models do not require document-summary paired data and instead they tackle a sentence-selection problem.
	TextRank \cite{mihalcea2004textrank} and LexRank \cite{erkan2004lexrank} encode sentences as nodes in a graph to select the most representative ones as a summary.
	\citet{zheng2019sentence} and \citet{rossiello2017centroid} advance upon TextRank and LexRank by using BERT \cite{devlin2018bert} to compute sentence similarity and replacing TF-IDF weights with word2vec embeddings respectively.
	In abstractive summarization, some approaches focus on learning unsupervised sentence compression with small-scale texts \cite{fevry2018unsupervised,baziotis2019seq,west2019bottlesum}, while TED \cite{yang2020ted} proposes a transformer-based architecture with pretraining on large-scale data.
	MeanSum \cite{chu2018meansum} generates a multi-document summary by decoding the average encoding of the input texts, where the autoencoder and the summarization module are interactive.
	\citet{bravzinskas2019unsupervised} and \citet{amplayo2020unsupervised} extend MeanSum by using a hierarchical variational autoencoder and denoising a noised synthetic dataset.
	However, none of these methods accommodate the multi-speaker scenario in dialogues.
	\begin{table*}[t!]
		\centering
		\small
		\begin{tabular}{c|ccc|ccc|ccc|ccc}
			\toprule
			\multirow{3}{*}{Model} &
			\multicolumn{6}{c|}{MultiWOZ} &
			\multicolumn{6}{c}{Taskmaster}  \\
			&\multicolumn{3}{c|}{Customer} & \multicolumn{3}{c|}{Agent}  &\multicolumn{3}{c|}{Customer} & \multicolumn{3}{c}{Agent}\\
			& R-1 & R-2 & R-L & R-1 & R-2 & R-L & R-1 & R-2 & R-L & R-1 & R-2 & R-L \\
			\midrule
			LexRank & 23.54 & 2.63 & 13.43 & 24.35 & 2.79 & 13.29 & 21.64 & 1.83 & 12.86 & 21.54 & 1.90 & 12.15 \\
			Word2Vec & 23.80 & 2.96 & 13.37 & 24.15 & 2.72 & 13.92 & 21.43 & 2.03 & 12.32 & 21.57 & 2.07 & 12.46 \\
			\midrule
			MeanSum & 25.93 & 4.42 & 14.52 & 26.49 & 4.49 & 15.43 & 24.01 & 3.31 & 13.55 & 24.08 & 3.24 & 14.31 \\
			Copycat & 26.86 & 4.81 & 16.35 & 26.92 & 4.37 & 16.12 & 24.86 & 4.23 & 14.81 & 25.05 & 3.71 & 15.19 \\
			VAE & 26.08 & 4.25 & 14.84 & 26.80 & 3.76 & 15.27 & 24.29 & 3.15 & 14.40 & 24.99 & 3.29 & 14.35 \\
			\midrule
			SuTaT-LSTM & \textbf{28.51} & \textbf{5.60} & \textbf{17.20} & \textbf{28.71} & \textbf{5.67} & \textbf{17.49} & \textbf{26.61} & \textbf{4.89} & \textbf{16.09} & \textbf{26.67} & \textbf{4.80} & \textbf{15.74} \\
			SuTaT-Tran & 26.82 & 4.80 & 16.08 & 27.11 & 4.88 & 15.52 & 25.20 & 3.98 & 15.33 & 25.19 & 4.12 & 14.81 \\
			\midrule
			\multicolumn{12}{c}{\textit{Ablation Study (with LSTM Encoders and Decoders)}}\\
			\midrule
			SuTaT w/o LS & 24.78 & 3.55 & 14.08 & 25.11 & 4.09 & 14.16 & 23.05 & 3.05 & 13.00 & 23.41 & 3.15 & 13.12 \\
			SuTaT w/o Att & 26.69 & 5.00 & 15.59 & 27.00 & 5.26 & 15.97 & 25.08 & 4.26 & 14.65 & 25.25 & 4.28 & 14.93 \\
			SuTaT w/o copy & 27.65 & 5.23 & 16.01 & 27.67 & 5.47 & 16.42 & 25.28 & 4.80 & 14.97 & 25.15 & 4.47 & 15.16 \\
			\bottomrule
		\end{tabular}
		\caption{ROUGE scores on the MultiWOZ and Taskmaster test sets.}
		\label{tab:rouge}
	\end{table*}
	
	\section{Experimental Details}
	
	We perform experiments with two variants of SuTaT: one equipped with LSTM encoders and decoders (SuTaT-LSTM), and the other equipped with Transformer encoders and decoders (SuTaT-Tran).
	
	\subsection{Dataset}
	The experiments are conducted on two dialogue datasets: MultiWOZ-2.0 \cite{budzianowski2018multiwoz} and Taskmaster-1 \cite{byrne2019taskmaster}.
	MultiWOZ consists of $10438$ goal-oriented human-human written dialogues between customers and agents, spanning over $7$ domains such as booking hotels, booking taxis, etc.
	$3406$ of them are single-label and $7302$ of them are multi-label.
	In the experiment, we split the dataset into $8438$, $1000$, and $1000$ dialogues for training, testing, and validation.
	Taskmaster consists of $13215$ goal-oriented dialogues, including $5507$ spoken and $7708$ written dialogues.
	In this work we only use the written dialogues which is created by human workers based on scenarios outlined for one of the six tasks, such as ordering pizza, ordering movie tickets, etc.
	The dataset is split into $6168$, $770$, and $770$ dialogues for training, testing, and validation.
	
	\subsection{Baselines}
	To validate the effectiveness of SuTaT, we compare the two variants against the following baselines: unsupervised extractive summarization methods LexRank \cite{erkan2004lexrank} and Word2Vec \cite{rossiello2017centroid}; unsupervised abstractive summarization methods MeanSum \cite{chu2018meansum} and Copycat \cite{bravzinskas2019unsupervised}.
	In addition, we train a vanilla text VAE model \cite{bowman2015generating} with our unsupervised summarization module as another baseline.
	
	Since we are the first work that summarizes for each speaker in a dialogue, some modifications need to be made on baselines to make fair comparisons with our model.
	To make the unsupervised summarization baseline models adapt to the two-speaker scenario in tete-a-tetes, we train two models for each baseline with either customer utterances or agent utterances.
	During testing, the customer summaries and agent summaries are generated by the two trained models of each baseline, which are used either separately for automatic and human evaluation or concatenated together for the classification experiment.
	
	\subsection{Settings}
	We fine-tune the parameters of SuTaT on the validation set.
	VAE-based text generative models can suffer from posterior collapse where the model learns to ignore the latent variable \cite{bowman2015generating}.
	We employ KL-term annealing and dropping out words during decoding to avoid posterior collapse.
	For KL annealing, the initial weights of the KL terms are 0, and then we gradually increase the weights as training progresses, until they reach the KL threshold of 0.8; the rate of this increase is set to 0.5 with respect to the total number of batches.
	The word dropout rate during decoding is 0.4.
	The latent variable size is 300 for both customer and agent latent variables.
	$\alpha$ that controls weights of two objective functions in Equation \ref{eq:obj} is set to 0.4.
	The word embedding size is 300.
	For the bidirectional LSTM encoder and LSTM decoder, the number of hidden layers is 1 and the hidden unit size is 600.
	For the Transformer encoder and decoder, the number of hidden layers is 1 and the number of heads in the multi-head attention is set to 10.
	The number of heads in the sentence-level self-attention is also 10.
	The hidden unit size of the MLPs in $p(\zv_y|\zv_x)$ is 600.
	The annealing parameter $\tau$ for soft-argmax in Equation \ref{eq:soft} is set to $0.01$.
	During training, the learning rate is 0.0005, the batch size is 16, and the maximum number of epoch is 10.
	SuTaT is implemented in pytorch and trained using a NVIDIA Tesla V100 GPU with 16GB.
	
	\begin{table*}[t!]
		\centering
		\small
		\begin{tabular}{c|cc|cc|cc|cc}
			\toprule
			\multirow{3}{*}{Model} &
			\multicolumn{4}{c|}{MultiWOZ} &
			\multicolumn{4}{c}{Taskmaster}  \\
			&\multicolumn{2}{c|}{Customer} & \multicolumn{2}{c|}{Agent}  &\multicolumn{2}{c|}{Customer} & \multicolumn{2}{c}{Agent}\\
			& PPL & KL & PPL & KL & PPL & KL & PPL & KL \\
			\midrule
			MeanSum & 3.58 & - & 3.65 & - & 5.57 & - & 5.48 & - \\
			Copycat & 3.46 & 0.75 & 3.42 & 0.73 & 5.41 & 0.96 & 5.23 & 0.93 \\
			VAE & 3.64 & 0.50 & 3.59 & 0.48 & 5.63 & 0.63 & 5.75 & 0.66 \\
			\midrule
			SuTaT-LSTM & 3.27 & 0.79 & 3.39 & 0.82 & 5.31 & 1.02 & 4.56 & 0.88 \\
			SuTaT-Tran & \textbf{1.77} & 0.28 & \textbf{2.10} & 0.34 & \textbf{2.48} & 0.35 & \textbf{2.52} & 0.36 \\
			\bottomrule
		\end{tabular}
		\caption{Language modeling results on MultiWOZ and Taskmaster. Lower is better for PPL.}
		\label{tab:lm}
	\end{table*}

	\subsection{Reference Summaries}
	In this work, we define the dialogue summary as summarizing for each speaker in a dialogue and there is no such annotated dataset available.
	To validate the effectiveness of SuTaT and compare with baselines, we follow the setting in \cite{chu2018meansum} to collect $200$ abstractive summaries for a subset of each dataset.
	Workers were presented with $10$ dialogues from MultiWOZ and $10$ dialogues from Taskmaster and asked to write summaries that ``best summarize both the content and the sentiment for each speaker''.
	We asked workers to ``write your summaries as if your were the speaker (e.g. `I want to book a hotel.' instead of `The customer wants to book a hotel.') and keep the length of the summary no more than one sentence''.
	The collected summaries are only used as reference summaries for testing and not used for model-tuning.
	These reference summaries cover all domains in both datasets and will be released later.
	
	\section{Results}
	We conduct the majority of experiments to show the superiority of SuTaT on unsupervised dialogue summarization.
	We use the labeled reference summaries for ROUGE-score-based automatic evaluation and human evaluation to compare with baseline methods.
	We further demonstrate the effectiveness of SuTaT by analyzing the language modeling results and using generated summaries to perform dialogue classification.
	In addition, we show that SuTaT is capable of single-turn conversation generation.

	\subsection{Unsupervised Dialogue Summarization}
	
	\begin{table}[t!]
		\centering
		\small
		\begin{tabular}{c|ccc|ccc}
			\toprule
			\multirow{2}{*}{Model} &
			\multicolumn{3}{c|}{MultiWOZ} &
			\multicolumn{3}{c}{Taskmaster} \\
			& Info & Read & Corr & Info & Read & Corr \\
			\midrule
			Reference & 5.43 & 4.73 & 4.52 & 5.39 & 4.57 & 4.60 \\
			\midrule
			MeanSum & 2.57 & 3.15 & 2.64 & 2.98 & 3.29 & 3.05 \\
			Copycat & 2.89 & 3.37 & 3.00 & 3.04 & 3.49 & 3.07 \\
			VAE & 2.96 & 3.04 & 2.44 & 2.97 & 2.92 & 2.45 \\
			\midrule		
			\tiny{SuTaT-LSTM} & \textbf{3.68} & 3.48 & \textbf{4.25} & \textbf{3.61} & \textbf{3.53} & \textbf{4.20} \\
			\tiny{SuTaT-Tran} & 3.47 & \textbf{3.56} & 4.15 & 3.33 & 3.52 & 3.96 \\
			\bottomrule
		\end{tabular}
		\caption{Human evaluation results on informativeness, readability, and correlation of generated summaries.}
		\label{tab:human}
	\end{table}
	
	\paragraph{Automatic Evaluation}
	ROUGE \cite{lin-2004-rouge} is a standard summarization metric to measure the surface word alignment between a generated summary and the reference summary.
	In the experiments, we use ROUGE-1, ROUGE-2, and ROUGE-L to measure the word-overlap, bigram-overlap, and longest common sequence respectively.
	Table \ref{tab:rouge} shows the ROUGE scores for two SuTaT variants and the baselines.
	As we can see, our proposed SuTaT with LSTM encoders and decoders outperforms all other baselines on both datasets.
	SuTaT-LSTM performs better than SuTaT-Transformer on ROUGE scores, the reason could be that Transformer decoders are too strong so the encoders are weakened during training.
	In general, the unsupervised abstractive models perform better than unsupervised extractive models.
	Compared with other unsupervised abstractive summarization baselines equipped with LSTM encoders and decoders, SuTaT-LSTM has a big performance improvement.
	We believe this is because SuTaT accommodates the two-speaker scenario in tete-a-tetes so that the utterances from each speaker and their correlations are better modeled.
	
	In addition, we evaluate reconstruction performances of the language modeling based methods with perplexity (PPL), and check the posterior collapse for the VAE-based methods with KL divergence.
	The results for MultiWOZ and Taskmaster are shown in Table \ref{tab:lm}.
	As can be seen, SuTaT-Tran has much better PPL scores than other competing methods on both datasets, showing the transformer decoders are effective at reconstructing sentences.
	Consequently, due to the powerful decoders, SuTaT-Tran has smaller KL divergences which can lead to posterior collapse where the encoders tend to be ignored.

	\paragraph{Human Evaluation}
	Human evaluation for the generated summaries is conducted to quantify the qualitative results of each model.
	We sample 50 dialogues that are labeled with reference summaries from the MultiWOZ and taskmaster test set (25 each).
	With the sampled dialogues, summaries are generated from the unsupervised abstractive approaches: MeanSum, Copycat, VAE, SuTaT-LSTM, and SuTaT-Tran.
	We recruit three workers to rank the generated summaries and reference summaries from 6 (the best) to 1 (the worst) based on three criteria: Informativeness: a summary should present the main points of the dialogue in a concise version; Readability: a summary should be grammatically correct and well structured; Correlation: the customer summary should be correlated to the agent summary in the same dialogue.
	
	The average ranking scores are shown in Table \ref{tab:human}.
	As we can see, SuTaT-LSTM achieves the best informativeness and correlation results on both datasets while SuTaT-Tran also has good performances, further demonstrating the ability of SuTaT on generating informative and coherent dialogue summaries.
	In general, the two SuTaT models have better human evaluation scores than baseline models, especially on correlation scores where the results are close to reference summaries.
	This is because SuTaT exploits the dependencies between the customer latent space and the agent latent space, which results in generating more correlated customer summaries and agent summaries.
	
	\paragraph{Ablation Study}
	We perform ablations to validate each component of SuTaT by: removing the variational latent spaces (SuTaT w/o LS) so the encoded utterances are directly used for embedding, removing the sentence-level self-attention mechanism (SuTaT w/o Att), and removing the partial copy mechanism (SuTaT w/o copy).
	We use LSTM encoders and decoders for all ablation models.
	The results for ablation study in Table \ref{tab:rouge} show that all the removed components play a role in SuTaT.
	Removing the latent spaces has the biggest influence on the summarization performance, indicating that the variational latent space is necessary to support our design which makes the agent latent variable dependent on the customer latent variable.
	The performance drop after removing the sentence-level self-attention mechanism shows that using weighted combined utterance embedding is better than simply taking the mean of encoded utterances.
	Removing the partial copy has the smallest quality drop.
	However, taking the dialogue example in Table \ref{tab:example}, without the partial copy mechanism SuTaT can generate the following summaries:
	\begin{quote}
		\textbf{Customer Summary}: i would like to book a hotel in cambridge on tuesday .
		
		\textbf{Agent Summary}: i have booked you a hotel . the reference number is lzludtvi . can i help you with anything else ?
	\end{quote}
	The generated summaries are the same except for the wrong reference number which is crucial information in this summary.
	
	\subsection{Classification with Summaries}
	\begin{table}[t!]
		\centering
		\small
		\begin{tabular}{c|c|c}
			\toprule
			Model & MultiWOZ & Taskmaster \\
			\midrule
			MeanSum & 0.76 & 0.70 \\
			Copycat & 0.77 & 0.72 \\
			VAE & 0.66 & 0.62 \\
			\midrule		
			SuTaT (unsupervised) & 0.85 & 0.79 \\
			SuTaT (supervised) & 0.99 & 0.96 \\
			\bottomrule
		\end{tabular}
		\caption{AUC scores for domain classfication with generated summaries, where MultiWOZ is multi-label and Taskmaster is single-label.}
		\label{tab:auc}
	\end{table}
	
	A good dialogue summary should reflect the key points of the utterances.
	We perform dialogue classification based on dialogue domains to test the validity of generated summaries.
	First we encode the generated customer summary and agent summary into $\ev_X$ and $\ev_Y$ using the trained encoders of each model, which are then concatenated as features of the dialogue for classification.
	In this way, the dialogue features are obtained unsupervisedly.
	Then we train a separate linear classifier on top of the encoded summaries.
	We use SuTaT with LSTM encoders and decoders for this task.
	As shown in Table \ref{tab:auc}, SuTaT outperforms other baselines on dialogue classification, indicating the SuTaT generated summaries have better comprehension of domain information in the dialogue.
	
	We can also perform supervised classification by using
	$\sv_X$ and $\sv_Y$ from SuTaT as features to train a linear classifier.
	The cross entropy loss is combined with Equation \ref{eq:obj} as the new objective function where all parameters are jointly optimized.
	As can be seen in Table \ref{tab:auc}, the supervised classification results are as high as $0.99$ on MultiWOZ and $0.96$ on Taskmaster, further demonstrating the effectiveness of SuTaT.

	\subsection{Single-Turn Conversation Generation}
	
	The design of the conditional generative module in SuTaT enables generating novel single-turn conversations.
	By sampling the customer latent variable from the standard Gaussian $\zv_x\sim\Ncal(\mathbf{0},\Imat)$ and then sampling the agent latent variable $\zv_y\sim p(\zv_y|\zv_x)$, SuTaT can produce realistic-looking novel dialogue pairs using the customer decoder and agent decoder.
	Table \ref{tab:example2} shows three examples of novel single-turn conversations generated by SuTaT using randomly sampled latent variables.
	We can see that the dialogue pairs are closely correlated, meaning the dependencies between two latent spaces are successfully captured.
	
	\begin{table}
		\centering
		\def\arraystretch{1.0}
		\setlength{\tabcolsep}{3pt}
		\small
		\begin{tabular} {p{0.55in} p{2.25in}}
			\toprule
			\textbf{Customer:} & yes , yes . are there any multiple sports places that i can visit in ?\\ 
			\textbf{Agent:} & sorry , there are none locations in the center of town . would you like a different area ?\\
			\midrule
			\textbf{Customer:} & yes please . book for the same group of people at 13:45 on thursday .\\ 
			\textbf{Agent:} & your booking was successful and your reference number is minorhoq .\\
			\midrule
			\textbf{Customer:} & hi , i am looking for a place to stay . the west should be cheap and doesn't need to have internet .\\ 
			\textbf{Agent:} & there are no hotels in the moderate price range . would you care to expand other criteria ?\\
			\bottomrule
		\end{tabular}
		\caption{Examples of single-turn conversations generated by the conditional generative module of SuTaT.}
		\label{tab:example2} 
	\end{table}
	
	\section{Conclusion}
	We propose SuTaT, an unsupervised abstractive dialogue summarization model, accommodating the two-speaker scenario in tete-a-tetes and summarizing them without using any data annotations.
	The conditional generative module models the customer utterances and agent utterances separately using two encoders and two decoders while retaining their correlations in the variational latent spaces.
	In the unsupervised summarization module, a sentence-level self-attention mechanism is used to highlight more informative utterances.
	The summary representations containing key information of the dialogue are decoded using the same decoders from the conditional generative module, with the help of a partial copy mechanism, to generate a customer summary and an agent summary.
	The experimental results show the superiority of SuTaT for unsupervised dialogue summarization and the capability for more dialogue tasks.
	
	\bibliography{main}
	\bibliographystyle{acl_natbib}
	
\end{document}